# Process Discovery using Classification Tree Hidden Semi-Markov Model

Yihuang Kang
*Department of Information Management*
*National Sun Yat-sen University*
*ykang@mis.nsysu.edu.tw*

Vladimir Zadorozhny
*School of Information Sciences*
*University of Pittsburgh*
*vladimir@sis.pitt.edu*

**Abstract**

*Various and ubiquitous information systems are being used in monitoring, exchanging, and collecting information. These systems are generating massive amount of event sequence logs that may help us understand underlying phenomenon. By analyzing these logs, we can learn process models that describe system procedures, predict the development of the system, or check whether the changes are expected. In this paper, we consider a novel technique that models these sequences of events in temporal-probabilistic manners. Specifically, we propose a probabilistic process model that combines hidden semi-Markov model and classification trees learning. Our experimental result shows that the proposed approach can answer a kind of question–"what are the most frequent sequence of system dynamics relevant to a given sequence of observable events?". For example, "Given a series of medical treatments, what are the most relevant patients' health condition pattern changes at different times?"*

**Keywords:** Hidden Semi-Markov Models, Classification and Regression Tree, Process Discovery, Temporal Data Mining

## 1. Introduction

The explosion of ubiquitous information systems has resulted in the exponential growth of operational event log data, which also has introduced the new age of the "Big Data" [1]. For example, Bedside Medical Device Interfaces provide a set of tools that automatically logs information from devices at the patient's bedside on Intensive Care Unit (ICU) monitors; a disease outbreak detection system records the numbers of outpatient visits for some particular diseases; and wireless sensor networks, such as air pollution monitoring and sea surface temperature detection systems, are deployed in an area to keep detecting and recording changes of physical or environmental conditions. These logs, however, are often used for monitoring purposes and are rarely created for further data analyses, such as underlying procedure discovery and business process auditing. By tracking the dynamics of the patterns identified by domain experts and/or pattern classification techniques from these logs over time, we can understand the phenomena of interests and how they evolve.

We define the *process* as *"a series of activities or state transitions of a dynamic system that produce some specific, either deterministic or probabilistic, outcomes."* [2]. The *process discovery* refers to a series of actions that collect activities of instances (e.g. customers of an online retail store) or changes of system states from event logs of information systems; use the logs to build process models that best describe the patterns; track the development of the systems by monitoring the changes of patterns; and evaluate the conformance of the development with expected models. Many approaches related to the process discovery have been proposed in different fields, such as process mining [3] and workflow management system [4]. These approaches aim at investigating how to use event logs of multiple instances to discover the underlying process models represented as the workflows and check whether the processes conform to expected process models. In this paper, however, we consider probabilistic process models that work on temporal sequence data. The temporal sequence refers to discrete event sequence data with event durations. Consider a simple example that we observe a weather system to understand the connection between the weather condition and temperature in terms of feeling. We can discretize the temperature in degrees centigrade C, (C < 15, 15 ≤ C < 25, 25 ≤ C) into three different states (Cold, Warm, and Hot). Also, we here only consider three possible weather conditions (Sunny, Cloudy, and Rainy) as observations. Suppose we observe a sequence of weather conditions, "Sunny → Cloudy → Rainy", most relevant temperature state sequences could be "Hot → Warm → Cold", "Warm → Warm → Cold", or any other permutations of three temperature states. We are interested in developing process models that find such patterns—"the most probable sequence of states given a sequence of observations".

In Figure 1, we illustrate and define the problems we cope with. Assume that we are monitoring a discrete dynamic system by collecting event log data that provides information about conditions of the system. We are interested in changes of well pre-defined observations of a phenomenon, (e.g. the weather conditions of aforementioned weather system), and we have some information about the system that may contribute to the changes.

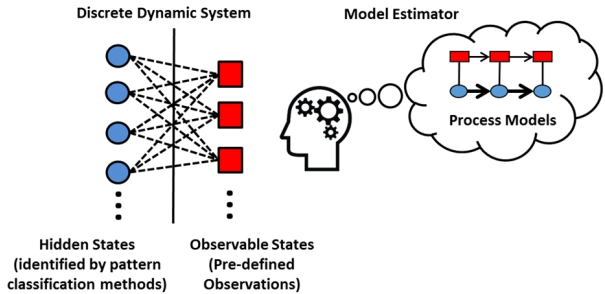

**Figure 1. Process modeling with discrete dynamic systems**

Figure 1 shows that we can directly see the pre-defined observable states (also called "observations"). On the other hand, there are some hidden states of the system that have impact on the changes of the observations. These hidden states may be identified by applying data discretization or pattern classification methods. Provided that there are some connections between the observations and hidden states, we can build the probabilistic process models that discover hidden state patterns given sequences of observations so as to understand and infer the development of the system. Therefore, we need techniques that help us a) identify hidden states (predictors) related to the changes of the observation (outcome); b) determine appropriate number of hidden states that can also be easily interpreted by human; c) create sequences of the hidden states with durations most relevant to the sequence of observations.

The rest of this paper is organized as follows. In Section 2, we review the backgrounds of the process model and discovery. Tree-based pattern classification methods and Markov models used in the proposed approach are also discussed. We consider the proposed probabilistic process models in Section 3. In Section 4, we present the experimental results, demonstrate how the proposed approach solves the problems, and discuss possible applications.

## 2. Background and related work

Process models are abstract models that help understand and describe underlying activities of a system from its event logs. As discussed in Section 1, many approaches [3–4] have been proposed to build deterministic process models. These process models are often represented by graphical notation languages, such as Petri Net, Business Process Management and Notation, and UML Activity Diagram [5]. In this paper, however, we consider building probabilistic models instead. Specifically, we assume the system we monitor is a *discrete dynamic system* [6] that can be represented by transitions of system states—sequence of states with durations and corresponding observations. As these state transitions could be probabilistic, we can model them using probabilistic modeling techniques, such as Markov models and Dynamic Bayesian Networks of Probabilistic Graphical Models [7]. In this paper, we use a specialization of the Markov Model—Hidden Markov model (HMM) [8].

Markov models are simple probabilistic models used to cope with the temporal sequences. They assume that the system is stochastic and with Markov property, which means the development of the dynamic system is assumed to be a random process that the current state of the system has the information about the next states. But, the next state only depends on the present state and are independent of past states. All the state transitions are probabilistic. Various kinds of Markov models are proposed and used in many real-world applications, such as Google PageRank [9] and speech recognitions [10]. Here, we consider using an extension of the HMM, *hidden semi-Markov models* (HSMM), which assumes that each hidden states has variable duration and may produce multiple observations while in the state.

As most event logs generated by aforementioned monitoring systems are not such temporal sequences, researchers in data mining communities have proposed using pattern classification [11] and time-series representation [12] techniques to discretize data into state-observation temporal sequences. The goal of these data discretization techniques is to keep the signatures (e.g. the distance measure) of the original data in transformed data space. That is, the distances among these transformed data stream are guaranteed to be similar to the distances in the original space. In this paper, we use classification tree of *Classification and Regression Tree* (CART) [13] to identify the state-observation sequences, as the states identified by tree models, compared to other linear or non-linear classification methods, can be easily interpreted. Figure 2 shows an example of two different classification models that divide the data spaces into three areas using Classification Tree and Support Vector Machine [14]. We consider the model that predicts the "Weather Condition" (Sunny, Cloudy, and Rainy as green, blue, and red dots respectively) with "Atmospheric Pressure (in hPa)" and "Temperature (in Fahrenheit)" as the predictors.

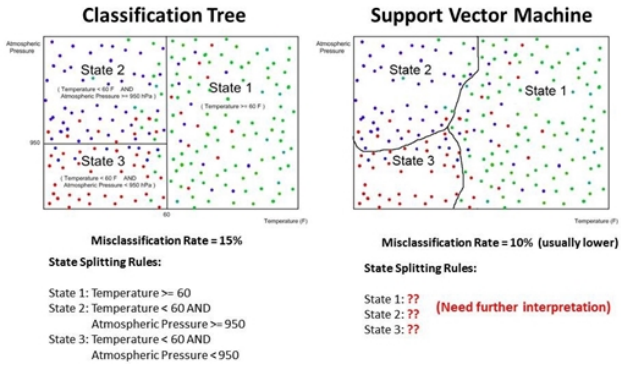

**Figure 2. Rules extraction from Classification Tree and SVM**

Depending on where the data points are in, each data point is assigned a state number that provides the information about the status of the weather system we monitor. For example, from the left tree model, we can learn that if the weather system is in State 1 (Temperature >= 60), most likely the weather condition is Sunny. Also, we can see that the misclassification rate of the SVM model is lower than the rate of the classification tree model, but the rules from the tree model are relatively easier to understand. Given the state splitting rules from the tree models, we can convert all the data points into state numbers and corresponding observations—the state-observation temporal sequences used to build the process models we discussed.

The state splitting methods (or pattern classification techniques) that define the states as patterns for HMM/HSMM is an active research area. In many applications of HMM, such as speech recognition, states have contextual and temporal domains. The number of states is initially one. Splits in both domains are tested and the best one is chosen. The grown HMM topology is retrained using *Baum-Welch algorithm* [10]. Maximum-Likelihood Successive-State-Splitting [15] is a well-known method that applies this algorithm to speech recognition fields. Instead, we propose using Classification Tree as the state splitting method, because states and the definitions of the states (IF-THEN rules from tree models) are easy-to-understand by humans. And, the state definitions can also provide us information about a system's state transitions.

An important issue about HMM is how to cope with variable state durations. In typical HMM, the state durations are assumed to be fixed. This is not natural for some dynamical systems that have different state sojourning times [16], i.e. the time intervals for such systems staying in each state are variable. In the recent decades, many researchers have indicated that modeling with HMM may be unrealistic and inaccurate when HMM is used in the applications that the state duration distributions (sojourning times) are different [17–18]. That is, the state durations are assumed to be all identical, which implies that the state durations are geometrically distributed. In this case, we consider the probability of spending continuous $m$ times/steps in $i$ state as:

$$d_i(m) = p_{ii}^{m-1}(1 - p_{ii})$$

where $d_i(m)$ is is the state duration (sojourning time) density and $p_{ii}$ is the probability that state $i$ transits to itself. We here model the probability that how many time ($m$ steps) a system will take to "leave" state $i$. That is, for example, Figure 3 shows a simulation of state duration distribution given $p_{ii} = 0.8$.

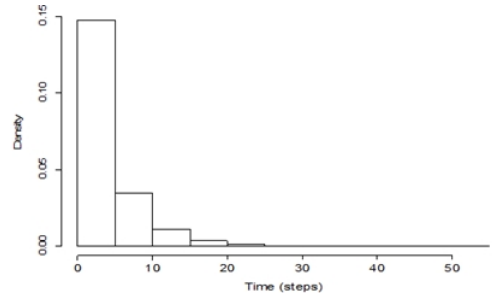

**Figure 3. State duration distribution with $p_{ii} = 0.8$**

We can see that, most likely, the system will stay in the state for less than 20 steps. Obviously, for most HMM applications, the state duration distributions are not necessarily in this particular "shape" (geometrically-distributed). In this paper, instead of using typical HMM, we therefore consider HSMM that explicitly models duration distribution for each state.

## 3. Proposed approach—Classification Tree Hidden Semi-Markov Model

In this section, we consider the proposed approach, Classification Tree Hidden Semi-Markov Model (CTHSMM), along with previous weather system as the example.

### 3.1 Process Discovery using CTHSMM

Suppose that we have a classification tree with three leaf nodes as shown in Figure 4. The tree model has divided the data space into three regions with predicted probabilities for three weather conditions (i.e. Sunny, Cloudy, and Rainy). We can then build an HSMM with three hidden states and an emission (observation) matrix from the predicted probabilities in three leaf nodes.

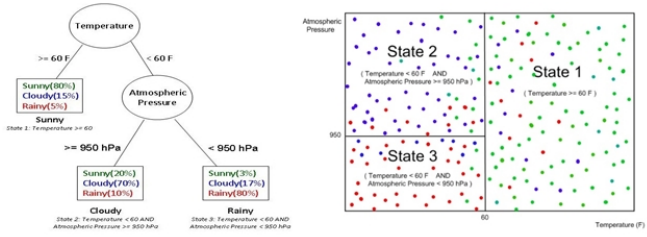

**Figure 4. A tree provides state splitting rules to divide data into 3 states/regions**

The state splitting rules extracted from the tree can then be used to convert each data point into a state number. By counting the number of state number transitions, we can create the state transition matrix of the HSMM. Table 1 shows an example of synthetic data with corresponding state numbers and their durations.

**Table 1. Weather system data with corresponding state numbers**

| Weather Condition | Temperature (F) | Atmospheric Pressure (hPa) | State Number | Duration (Hour) |
|---|---|---|---|---|
| Cloudy | 62 | 982 | 1 | 2 |
| Rainy | 50 | 950 | 2 | 1 |
| Rainy | 48 | 930 | 3 | 1 |
| Cloudy | 55 | 950 | 2 | 2 |
| Sunny | 65 | 980 | 1 | 6 |

As discussed in Section 2, the state durations are not necessarily identical in most real-world applications. The state duration distributions may also not be geometric if the weather system is unstable. In such cases, we should consider HSMM, as it explicitly estimates the duration distribution for each state [16]. Instead of using aforementioned Baum-Welch algorithm that iteratively re-estimates state transition probability matrix, we propose to estimate "in-state" (absorbing state) transition probabilities ($p_{ii}$) and "out-state" transition probabilities ($p_{ij}$, where $i \neq j$) separately. Figure 5 shows an example of the weather system with three states.

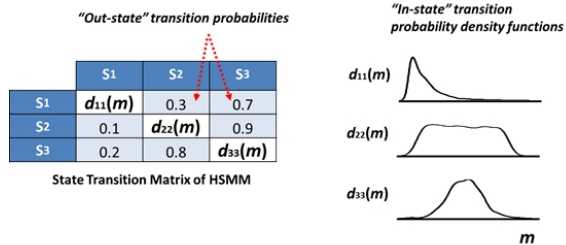

**Figure 5: State transition probability estimation of the HSMM**

The out-state transition probabilities can be computed from the relative frequencies of the out-state state number transitions from the raw data with identified presumable hidden state numbers. Note that diagonal cells of the left matrix in Figure 5 must be zero, as we estimate the in-state transition probabilities separately and do not consider these absorbing (sojourning) states. On the other hand, we can see that the state durations (sojourning time) could vary from different states and are assumed to be not necessarily geometrically-distributed. Also, as the hidden states are identified by the tree model, we may not have the information about the duration distribution for each state. Therefore, instead of making assumptions about the duration distributions, we propose to explicitly estimate the probability density function of the duration of each state from the training data by using Kernel Density Estimation with Gaussian kernel smoother and the rule-of-thumb estimation of the bandwidth [18] as defined:

$$h = \left(\frac{4\hat{\sigma}^5}{3n}\right)^{\frac{1}{5}}$$

where $n$ is the sample size and $\hat{\sigma}$ is the sample standard deviation. The estimated density functions for state durations are then used in the predictions of the Viterbi path [19] of HSMM for different lengths ($m$) of sojourning/absorbing states.

With the transition probability estimation shown in Figure 5 that addresses the problem of variable state duration distributions, the proposed approach is able to predict the most probable state transitions with variable time units (durations) given different lengths of observation sequences. To evaluate the accuracy of the Viterbi path prediction, we propose using a simple Hit Ratio and Longest Matched Run (LMRL) Ratio, which are ratios of numbers of matched states and longest matched state runs to total numbers of predicted states respectively. Figure 6 shows an example about how to calculate both ratios. Given that we have a sequence of weather conditions for total 18 hours, the most relevant hidden weather state transitions is S1 → S2 → S3 for 5, 4, and 9 hours respectively.

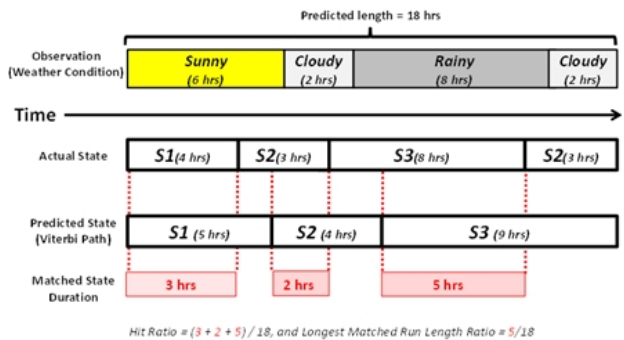

**Figure 6. Hit and LMRL Ratios for variable lengths of actual and predicted states**

Note that one tree model can create one particular CTHSMM. For any given dataset, we may build many tree models and then create multiple CTHSMMs with different parameters (i.e., number of hidden states, emission matrices, and state transition matrices). Therefore, the important issue needed to be addressed here is "which CTHSMM is the best?". We consider CTHSMM model selection in the next section.

## 3.2 Maximum Mutual Information Estimation of CTHSMM Parameters

As classification tree of CART is used in the proposed approach to build HSMM, we could consider techniques used in typical tree model selection. The most common techniques are tree pre-pruning and post-pruning. The pre-pruning is to halt the tree construction early by setting a threshold on a tree growth parameters (e.g. the minimum number of data points in a leaf node), whereas post-pruning is to prune subtrees from a grown tree. It is difficult to choose an appropriate tree pre-pruning threshold, as high thresholds may create oversimplified trees, whereas low thresholds may result in complicated and deep trees. The post-pruning method, on the other hand, employs cost-complexity pruning [13] algorithm that finds the balance between tree splitting cost and tree complexity. A tree with too few leaves creates a trivial CTHSMM, while a tree with too many leaves results in an CTHSMM with many states and complex state rules difficult to understand. Classification tree models are typically selected by both pre- and post-pruning methods and are evaluated by misclassification rate using $k$-fold cross validation (CV-MR) [13]. Here, however, we are not certain that good tree models will result in better HSMM. To evaluate an HSMM, we must consider one of its important parameter—observation (emission) matrix, which indicates how well the connection between the hidden states and observations. Take previous weather system as the example again. Assume that we have an HSMM from a tree model as shown in Figure 7.

|  | Sunny | Cloudy | Rainy |
|---|---|---|---|
| State 1 | 0.7 | 0.2 | 0.1 |
| State 2 | 0.1 | 0.1 | 0.8 |

Matrix X

|  | Sunny | Cloudy | Rainy |
|---|---|---|---|
| State 1 | 0.7 | 0.2 | 0.1 |
| State 2 | 0.1 | 0.1 | 0.8 |
| State 3 | 0.33 | 0.33 | 0.34 |

Matrix Y

|  | Sunny | Cloudy | Rainy |
|---|---|---|---|
| State 1 | 0.7 | 0.2 | 0.1 |
| State 2 | 0.1 | 0.1 | 0.8 |
| State 3 | 0.005 | 0.99 | 0.005 |

Matrix Z

**Figure 7. An observation matrix with one more different splits**

Suppose observation Matrix X is from a tree with two leaf nodes. We would like to have one more split to obtain one more state/leaf. A new tree with Matrix Z provides us more information than a tree with Matrix Y does, as we know if the system is in State 3, most likely the weather condition is Cloudy.

We can measure how much information an observation matrix can give us by calculating its *Mutual Information* (MI) [20] in bits, as defined:

$$MI = \sum_{i=1}^{o} \sum_{j=1}^{s} P(O_i, S_j) \log_2 \left( \frac{P(O_i, S_j)}{P(O_i) P(S_j)} \right)$$

where $P(O_i, S_j)$ are State-to-Observation joint probabilities, which can be obtained by multiplying the prior probabilities of the states $P(S_j)$ as defined by Bayes' theorem, i.e.

$$P(O_i, S_j) = P(O_i | S_j) P(S_j)$$

The MIs for Matrix X, Y, and Z are 0.4184, 0.3090, and 0.8312 bits respectively, which also suggests that MI is an appropriate measure to evaluate the observation matrices. Therefore, instead of selecting tree models based on the $k$-fold cross validation misclassification rate, we could consider a tree model that creates an HSMM with maximum MI—*Maximum Mutual Information Estimation* (MMIE) [20]. Specifically, here we consider a tree model with maximum MI given parameter *minbucket*, a minimum numbers of data points in a leaf node to limit the tree growth, as defined:

$$minbucket_{maxMI} = \underset{minbucket \in [1, |train_{data}|]}{argmax} MI(minbucket, train_{data})$$

where $train_{data}$ and $|train_{data}|$ are the training dataset used to build the model and the number of total data points (records) in the dataset. The goal is to choose a minbucket (between 1 and $|train_{data}|$) that maximizes the objective function (MI) that computes the mutual information of a given CTHSMM. In the next section, we discuss the role of MI again in more detail when it is applied to the selection of the best CTHSMM.

## 4. Experimental Result and Discussion

We demonstrate the proposed CTHSMM with a real-world dataset. All experiment are implemented in R version 3.1.1[21] with package *rpart* [22] and *mhsmm* [23]. A series of steps to build the CTHSMM is illustrated in Figure 8. These steps can be summarized as follows: 1) Divide data into two different parts as training and testing datasets; 2) Use CART with both post-pruning (cost-complexity) and pre-pruning (MMIE) methods to learn candidate CTHSMMs from the training dataset; 3) Apply state splitting rules to the predictors of the testing dataset to obtain presumable state sequences of testing dataset; 4) Generate Viterbi paths (predicted hidden state sequences) given the observation sequences from testing dataset; 5) Calculate the averages of hit and LMRL ratios and create plots for model evaluation.

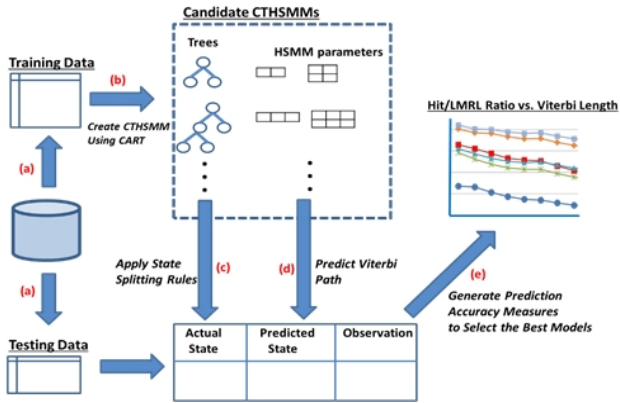

**Figure 8. CTHSMM model selection process**

The dataset used here is from the Children's Hospital of Pittsburgh (CHP). It consists of one categorical dependent variable as a patient's current location, in "ICU" or "Floor" (public ward), one continuous variable as the duration in hour, and five continuous independent variables as five vital signs for a patient: Diastolic Blood Pressure (DBP), Systolic Blood Pressure (SBP), Respiratory Rate per minute (RR), SpO2 Bedside Monitor (SPO2), and Body Temperature (Temp). There are 13,006 rows for total 359 patients who are children between one and six years old and were hospitalized in 2008. Table 2 shows a sample of CHP data for a patient. These patients have previously reported with respiratory problems. The goal of using this dataset is to explore whether CTHSMM can help doctors understand patients' vital sign pattern dynamics, evaluate patient respiratory complaint risk, and further improve hospital bed utilization rate.

**Table 2. A sample of CHP dataset**

| Diastolic Blood Pressure (mm Hg) | Systolic Blood Pressure (mm Hg) | Respiratory Rate (bpm) | SpO2 Bedside Monitor (%) | Temperature (C) | Location | Duration (Hour) |
|---|---|---|---|---|---|---|
| 64 | 117 | 29 | 100 | 37.5 | ICU | 1 |
| 65 | 110 | 21 | 99 | 37.5 | ICU | 1 |
| 65 | 110 | 21 | 99 | 37.5 | ICU | 1 |
| 65 | 110 | 21 | 98 | 37.5 | ICU | 1 |
| 66 | 90 | 26 | 96 | 36.7 | Floor | 2 |
| 67 | 97 | 27 | 98 | 36.7 | Floor | 1 |
| 67 | 97 | 27 | 96 | 36.7 | Floor | 2 |
| 65 | 94 | 26 | 98 | 37.1 | Floor | 2 |
| 65 | 94 | 26 | 98 | 37.1 | Floor | 3 |
| 68 | 87 | 15 | 98 | 37.3 | Floor | 1 |

We first divided CHP data by randomly sampling approximately 70% patients from CHP data as the training dataset and the rest of 30% data are considered as the testing dataset. There are total 8,734 and 4,272 rows for training and testing datasets respectively. The proposed steps of CTHSMM learning and selection process is then applied to the training dataset to construct candidate CTHSMMs. We first used the cost-complexity post-pruning algorithm to continuously prune a fully-growth tree model (with 1% of training data records as the threshold of the minimum number of data points in a leaf node). There are one fully-grown tree and six pruned tree models created after the post-pruning process. One additional candidate CTHSMM is also created by using the proposed MMIE. We found that, for the training dataset with 8,734 rows from CHP data, a tree model as shown in Figure 9 with minbucket = 101, can maximize the MI (0.2098 bits).

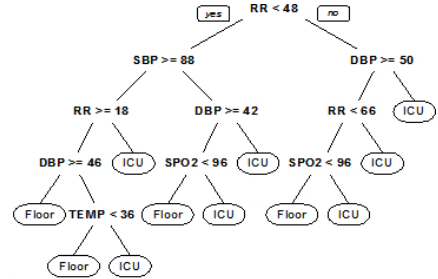

**Figure 9. Tree selected by MMIE, MI = 0.2098 bits, CV-MR = 0.1975**

Next, the testing dataset of CHP is used to evaluate these eight candidate CTHSMMs. We applied state splitting rules from each candidate CTHSMM to mapping predictors (five vital signs) into presumable hidden state sequences (vital sign pattern dynamics). On the other hand, each patient's location sequence (observation sequence) of the testing dataset is used to infer most probable state sequences (predicted Viterbi path) with different lengths (time periods). Comparing the actual and predict state sequences, we can then compute the aforementioned average hit and LMRL ratio. Figure 10 shows both measures for seven CTHSMMs with different lengths of predicted sequences/times in hour.

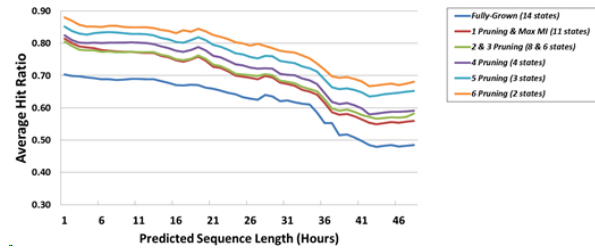

**(a)**

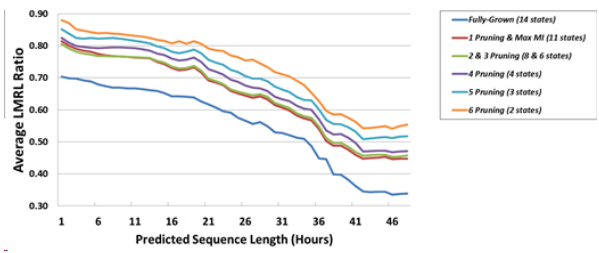

**(b)**

**Figure 10. Average Hit and LMRL Ratio with different hours of prediction**

It suggests that predicting longer time would result in lower accuracy in terms of average hit and LMRL ratios. Also, we expect that CTHSMMs with fewer states would inevitably perform better. For example, the CTHSMM with 6-pruning tree has highest hit and LMRL ratios, because there are only two hidden states and the chance of hits are relatively higher. However, the CTHSMM selected based on MMIE (11 hidden states) performs comparatively well even with more states compared to CTHSMMs with fully-grown or 2 & 3-pruning trees, which also suggest that CTHSMM selected by MMIE provides us more information and relatively higher accuracy of Viterbi path prediction.

Let's consider real-world applications of the proposed approach. Here, we choose the CTHSMM built from the tree in Figure 9 to infer most probable a patient's vital sign pattern dynamics given sequence of his/her location changes, because the model with maximum MI has shown that it could provide us some information (from the state splitting rules) and relatively better accuracy of sequence prediction. Table 3 shows its observation matrix with the state definition rules. Note that those vital signs out of normal ranges are colored in red.

**Table 3. Observation matrix with state definition rules for the CTHSMM with maximum MI**

| State | Floor | ICU | State Definition Rule |
|---|---|---|---|
| S1 | 0.0573 | 0.9427 | RR>=48 & DBP< 50 |
| S2 | 0.1103 | 0.8897 | RR< 18 & SBP>=88 |
| S3 | 0.1161 | 0.8839 | RR< 48 & SBP< 88 & DBP< 42 |
| S4 | 0.1608 | 0.8392 | RR>=66 & DBP>=50 |
| S5 | 0.8271 | 0.1729 | RR< 48 & RR>=18 & SBP>=88 & DBP>=46 |
| S6 | 0.6230 | 0.3770 | RR< 48 & SBP< 88 & DBP>=42 & SPO2< 96 |
| S7 | 0.3393 | 0.6607 | RR< 48 & SBP< 88 & DBP>=42 & SPO2>=96 |
| S8 | 0.5965 | 0.4035 | RR>=48 & RR< 66 & DBP>=50 & SPO2< 96 |
| S9 | 0.3918 | 0.6082 | RR>=48 & RR< 66 & DBP>=50 & SPO2>=96 |
| S10 | 0.6903 | 0.3097 | RR< 48 & RR>=18 & SBP>=88 & DBP< 46 & TEMP< 36 |
| S11 | 0.3898 | 0.6102 | RR< 48 & RR>=18 & SBP>=88 & DBP< 46 & TEMP>=36 |

Again, the CTHSMM can help answer the question —"what are the most probable sequence patterns relevant to a given sequence of observable events?". Here we enumerate three possible application scenarios:

Scenario 1: "Suppose that a patient stayed in ICU for a while. Then he was moved to Floor. Based on selected CTHSMM, what would be the most probable the patient's vital sign state dynamics given different lengths of times (sequences) he stayed in ICU and Floor?" As shown in Figure 10b, if we predict state sequences longer than 21 hours, the accuracy of prediction are lower than 70% in terms of average LMRL ratio. Thus, we here only discuss the cases within 21 hours in the following examples. In Figure 11, we consider the case in Scenario 1 that the patient was first in ICU and then was moved to Floor. We can see that there is a "transition state" (e.g. S10) before the patient was moved to Floor, which seems reasonable, as patients would most likely be moved from ICU to Floor only when their vital sign conditions are better.

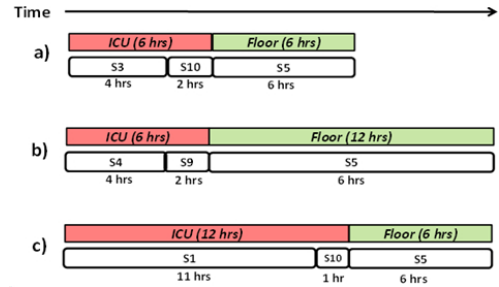

**Figure 11. Most probable vital sign sequences given different patients' location within 18 hours for Scenario 1**

Scenario 2: "Another situation is that a patient first stayed in Floor. Then his condition became worse. Doctors decided to move him to ICU and keep monitoring him. Given the selected CTHSMM, again, what would be the most probable the patient's vital sign patterns dynamics?" Here, we consider the reverse cases in Scenario 2 that the patient was first admitted, and then was moved from Floor to ICU. Figure 12 shows a similar situation that there is a transition state (S3) before a patient is moved to different location. It is intuitive that the doctors would move a patient to ICU most likely when the patient's is very bad (RR >= 48 and DBP < 50).

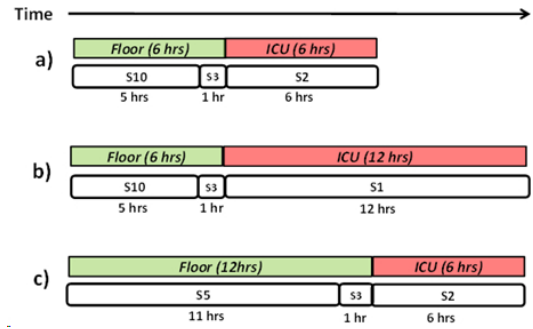

**Figure 12. Most probable vital sign sequences given different patients' location within 18 hours for Scenario 2**

Scenario 3: "Suppose that there is an unusual situation that a patient was admitted to the hospital but was moved between Floor and ICU back and forth several times, as his condition was never stable. We would like to know, in such cases, whether the patient vital signs would become stable." Scenario 3 is a dramatic case as there may be many observation transitions (location changes). Figure 13 shows sample results when we applied CTHSMM learned from the data to the prediction of most probable hidden state (vital sign pattern) sequences with different time periods.

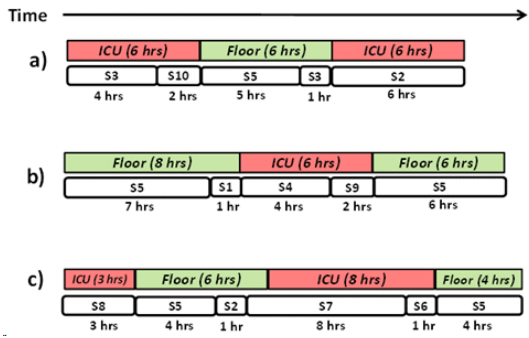

**Figure 13. Most probable vital sign sequences given different patients' location and times for Scenario 3**

We can see that, given different sequence of the observations (locations), the CTHSMM can both characterize more detail about state transitions with variable durations and discover most relevant sequences of hidden states (vital sign patterns).

## 5. Conclusion

We proposed a novel process discovery approach that integrates hidden semi-Markov model and classification trees to uncover hidden system state patterns and build a probabilistic process model simultaneously. The experimental results shows that it can help identify most frequent sequence of system state changes relevant to a given sequence of observable events. The state definition rules extracted from classification tree provides human-comprehensible information about the system dynamics. The most probable patterns of system state changes along with variable state durations can help decision maker to understand the system development in temporal-probabilistic manner.